\documentclass[conference]{IEEEtran}
\IEEEoverridecommandlockouts
\usepackage{cite}
\usepackage[bb=dsserif]{mathalpha}
\usepackage{amsmath,amssymb,amsfonts}
\usepackage{algorithmic}
\usepackage{graphicx}
\usepackage{textcomp}
\usepackage{comment}
\usepackage{booktabs}
\usepackage{xcolor}
\usepackage{url}
\usepackage{multirow}
\usepackage{comment}
\usepackage{makecell}
\usepackage{hyperref}
\usepackage{subcaption}
\usepackage{float}

\usepackage{bm}

\long\def\comment#1{}

\newcommand{\xv}{{\bf x}}

\renewcommand{\arg}{{\hbox{arg}}}

\def\BibTeX{{\rm B\kern-.05em{\sc i\kern-.025em b}\kern-.08em
    T\kern-.1667em\lower.7ex\hbox{E}\kern-.125emX}}
\begin{document}

\title{Robust Classification under Noisy Labels: A Geometry-Aware Reliability Framework for Foundation Models\\
% OTHER IDEAS:
%“Embedding Reliability for Noisy-Label Robustness in Foundation Models via Non-Negative Kernel Polytopes”
%“Enhancing Noisy-Label Robustness in Foundation Vision Models with NNK-Polytope Embedding Reliability”
%“Robust Classification under Noisy Labels: An NNK-Polytope Reliability Framework for Foundation Models”
%{\footnotesize \textsuperscript{*}\small\colorbox{yellow}{DEADLINE: JULY 15, 2025 \textbf{\url{https://camsap25.ig.umons.ac.be/call-for-papers.php}}}}
%\thanks{Papers cannot be longer than 5 pages (double-column IEEE template for conference proceedings), including all text, figures, and references. The 5th page cannot contain any text other than references.\\}
}
\author{\IEEEauthorblockN{Ecem Bozkurt}
\IEEEauthorblockA{\textit{Department of Electrical and Computer Engineering}\\
\textit{ University of Southern California}\\Los Angeles, CA, USA\\
bozkurt@usc.edu}
\and
\IEEEauthorblockN{Antonio Ortega}
\IEEEauthorblockA{\textit{Department of Electrical and Computer Engineering} \\
\textit{University of Southern California}\\
Los Angeles, CA, USA\\
aortega@usc.edu}}

\maketitle

\begin{abstract}

Foundation models (FMs) pretrained on large datasets have become fundamental for various downstream machine learning tasks, in particular in scenarios where obtaining perfectly labeled data is prohibitively expensive. In this paper, we assume an FM has to be fine-tuned with noisy data and present a two-stage framework to ensure robust classification in the presence of label noise without model retraining. Recent work has shown that simple k-nearest neighbor (kNN) approaches using an embedding derived from an FM can achieve good performance even in the presence of severe label noise. Our work is motivated by the fact that these methods make use of local geometry. In this paper, following a similar two-stage procedure, reliability estimation followed by reliability-weighted inference, we show that improved performance can be achieved by introducing geometry information. For a given instance, our proposed inference uses a local neighborhood of training data, obtained using the non-negative kernel (NNK) neighborhood construction. We propose several methods for reliability estimation that can rely less on distance and local neighborhood as the label noise increases. Our evaluation on CIFAR-10 and DermaMNIST shows that our methods improve robustness across various noise conditions, surpassing standard K-NN approaches and recent adaptive-neighborhood baselines.
\end{abstract}

\section{Introduction}
% Motivation: foundational vision models transforming vision tasks
% Noisy labels critical in safety-sensitive domains (healthcare, autonomous driving)
% Data collection and annotation are expensive and time-consuming
% Two-stage approach: (1) compute reliability score, (2) decision based on reliability
Foundation Models (FMs) are large-scale models pre-trained on large-scale datasets \cite{bommasani2021opportunities}. One of the key strengths of foundation models is their plug-and-play nature: once pre-trained, they can be applied directly without the need for additional parameter tuning and can adapt to a wide range of downstream tasks. 
%FMs are widely applied across various domains, especially in natural language processing (NLP) with models like BERT \cite{devlin2019bert}, GPT \cite{brown2020gpt3}, and T5 \cite{raffel2020t5}; in computer vision with models such as ViT \cite{radford2021Vit}, DINO \cite{oquab2023dinov2}, and SAM \cite{kirillov2023segment}; in multimodal applications with CLIP \cite{radford2021Vit} and Flamingo\cite{alayrac2022flamingo}; and in speech processing with models like Whisper \cite{radford2023whisper} and Wav2Vec 2.0 \cite{baevski2020wav2vec}.
%Modern foundational vision models %like CLIP and DINO 
%have revolutionized visual understanding by providing semantically rich embeddings. %how?
%ROBUST CLASSIFICATION 
%\section{Related Work}
% Previous methods
% k-NN confidence: neighbor agreement rate
% NNK graphs: capture local convex geometry
% WANN competitor: considers data homogeneity, k-NN neighborhood, LDA; ignores direct distances, similarity, geometry, density, outliers
% Clustering-based reliability: distances to centroids, k-means: one-center , nnk-means : multiple centers 
However, this ease of use comes with a downside: if the downstream task dataset contains corrupt labels or does not align well with the FM, retraining the model is not a feasible option because it is time-consuming and expensive to annotate all labels correctly. Therefore, it is important to identify alternatives to retraining that allow us to use FMs while taking into account label inaccuracy. 

%Corrupt labels have been a persistent problem in supervised learning settings; it is time-consuming and expensive to annotate all labels correctly. Not all downstream datasets are compatible with all foundation models. Since foundation models are large-scale, retraining them is not a feasible option. 

Current research on robust learning in the presence of label noise is generally divided into three different approaches: 1) sample selection, 2) loss adjustments, and 3) embedding space. \textit{Sample selection} methods focus on identifying and using clean (or likely clean) samples during training, while minimizing the impact of noisy ones. This approach can involve computationally intensive techniques, such as multi-network strategies or iterative filtering, as discussed in several studies \cite{han2018coteaching, jiang2018mentornet, song2019selfie, li2020dividemix, wei2020combating, xiao2022promix, zhang2024badlabel, fooladgar2024}. Sample selection methods typically struggle with handling \textit{rare} classes. In contrast, \textit{loss adjustment} methods use modified loss functions specifically designed to address label noise \cite{patrini2017losscorrection, tanaka2018jointopt, liu2020earlylearning, zhang2021understanding}. While these methods also have high computational complexity, there are lightweight alternatives available \cite{ghosh2017robust, zhang2018generalized, wang2019symmetry, ma2020normalized, zhou2021asymmetric, ye2024active}. Nevertheless, loss adjustment methods come with drawbacks, including limited interpretability, the risk of overfitting to incorrect labels, a need for large datasets, and challenges in managing ambiguous or incorrect labels.
%third approach is  ..... 

In this paper, we focus on approaches that use \textit{local geometry in the FM embedding space} to assess the reliability of each training sample for the downstream task. These approaches enhance classification robustness without necessitating retraining and offer greater interpretability.
%However, assessing the reliability is still a challenge, especially in safety-critical applications such as medical diagnosis and autonomous vehicles, where label noise can pose serious risks. 
%DIFFERENT STAGES
%Ongoing research focuses on developing robust learning and prediction schemes to address mislabeled classes. 
 %comment on fine-tuning option 
%
Di Salvo et al.~\cite{disalvo2025} introduced the Weighted Adaptive Nearest Neighbor (WANN) method for FMs, which enhances the traditional k-nearest neighbor (k-NN) classifier to tackle label noise in the embedding space. WANN dynamically adjusts the neighborhood size for each query based on local label consistency. Essentially, as label noise increases, larger neighborhoods can deliver more reliable decisions based on majority voting.  
WANN uses a two-stage pipeline:
(i) calculate a reliability score for each training data sample to reflect its trustworthiness, and 
(ii) utilize these scores in classification decisions for the test data to enhance robustness against noisy labels.
%, resulting in more robust predictions in noisy label conditions and providing greater interpretability without the need to retrain any model parameters.
%and LDA to deal with high-dim data 
%By contrast, standard k-NN treats every point equally and relies solely on majority voting, without capturing richer geometric structure. 
%Although WANN improves upon this by adapting neighborhood radii to data homogeneity, 
While demonstrating robustness to label noise, WANN has the limitation of using only k-NN to identify a neighborhood. In particular, WANN does not make use of distances in feature space; it cannot extend a neighborhood beyond a pre-specified value of $k$ neighbors, does not consider the relative position of neighbors, or local variations in density.
%Yet, WANN overlooks several important factors: the precise inter-feature distances, %alternative similarity metrics, 
%the geometry of the local embedding neighborhood, and local variations in the density. 
%and an explicit mechanism for identifying outliers versus truly ambiguous samples. 

In this paper, we follow the same pipeline as WANN. In the first stage, we propose \textit{local and global geometry-based approaches to estimate reliability}. 
%
%In this work, 
For local estimation, we use the non-negative kernel (NNK) \cite{Shekkizhar2019} graph construction, which results in a local, geometrically non-redundant neighborhood defining a polytope around each instance. We use the NNK weights, which quantify the relative similarity (proximity in feature space), and the size of the local polytope, along with label information, to estimate reliability for each sample in the training set. 
For global estimation, motivated by high label noise scenarios (where, since labels may be flipped, it is no longer possible to trust nearby instances), we use metrics based on supervised and unsupervised clustering. 
%distances become increasingly unreliable since labels can be leading to more refined measures of reliability. FMs provide high-dimensional embeddings, and as the level of label noise increases, the significance of the distances between the features diminishes. 
%To tackle this issue, rather than directly expanding the local neighborhood, 
%we employ a randomization technique called \textit{ensemble NNK reliability} to better characterize the reliability of the training data. In addition, 
%we explore clustering-based methods, such as k-means 
%and NNK-means \cite{shekkizhar2022nnkmeans}, 
%These methods allow us to include more global properties and determine the conditions under which they provide the best robust predictions in the presence of noisy labels.
In the \textit{inference stage}, each test instance is assigned a label based on a weighted estimate given the labels of its NNK neighbors. In high noise settings, the weights are a function of reliability only, while in low noise settings both NNK weights (based on relative distances) and reliability are combined. 

Our main contributions are as follows: 1) We use a novel, geometry‑aware neighborhood construction via the NNK algorithm in the embedding space,  
2) We introduce novel reliability metrics that leverage both distances and the shape of these local neighborhoods based on NNK weights and polytope diameter ratios,   
%(D/Dc) reliability
      %, and (iii) Ensemble NNK reliability that randomizes neighborhoods via repeated sub-sampling and averages results to boost robustness under high label noise.
3)  To further handle extreme noise or unfit data, we integrate global clustering based on k‑means %and NNK‑Means 
  for reliability estimation.  
%4) We develop flexible decision rules that integrate edge weights and reliability scores —either together or focusing solely on reliability— to adjust sensitivity based on feature distances.
%
We validate our methods on two vision tasks — CIFAR‑10 (a standard benchmark) and DermaMNIST (a challenging medical dataset) — under different noise types and levels, showing improvement over k‑NN and adaptive‑neighborhood baselines (ANN, WANN).

\section{Robust Classification Framework}
% Feature extraction: CLIP-ViT/32 and DINOv2-base
% Compute reliability metrics definitions
% Classification: reliability-weighted voting

\begin{figure}
    \centering
    \includegraphics[width=1\linewidth]{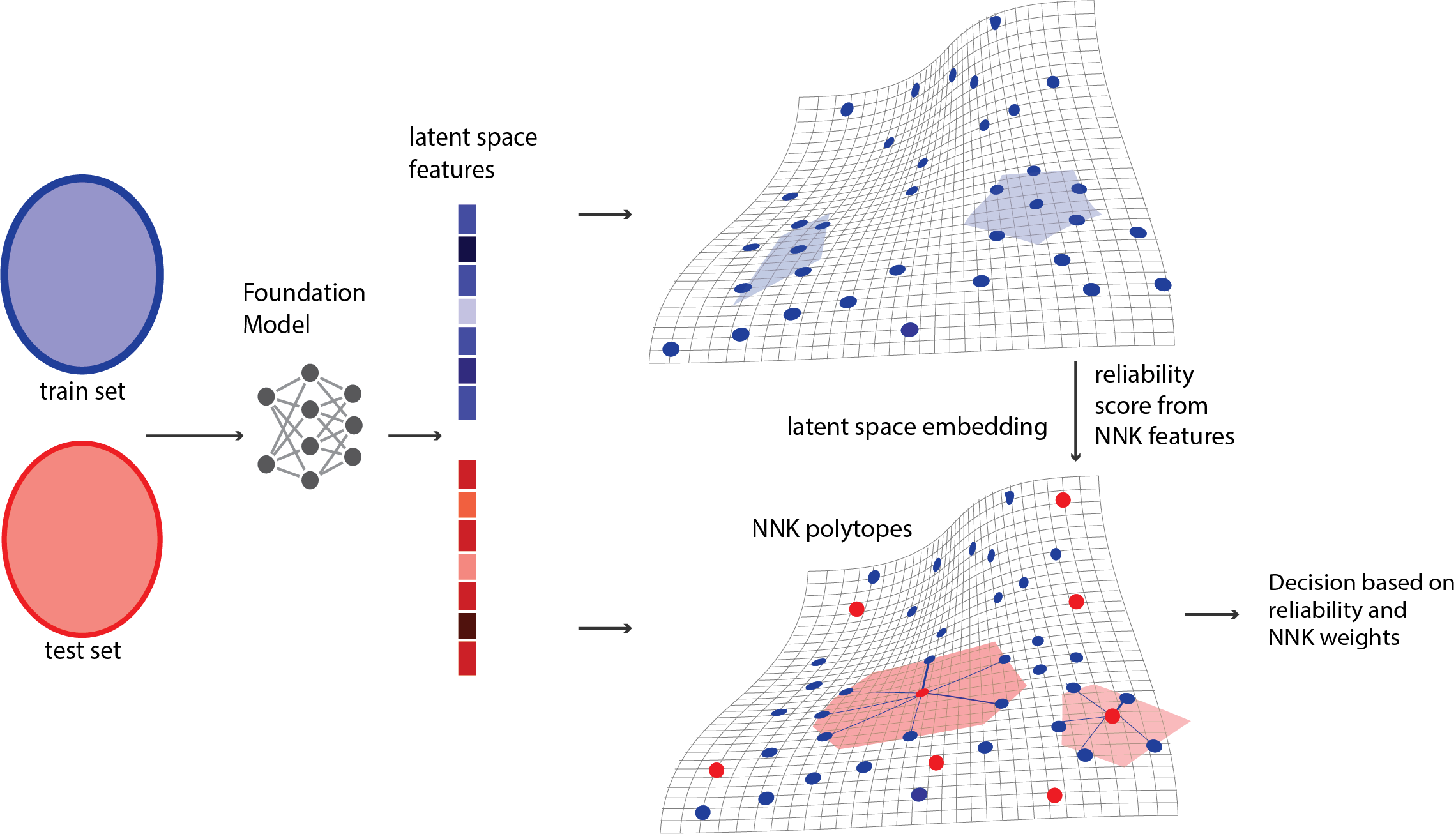}
    \caption{The latent space features for both the training and test datasets are obtained. The training dataset is utilized to estimate the reliability of each sample within. This reliability score is then applied to classify the test data using a reliability-weighted majority voting approach within the selected local neighborhood.}
    \label{fig:pipeline}
\end{figure}
%\subsection{Feature Extraction}
% Use CLIP-ViT/32 and DINOv2-base
% Resize images to 224×224 and normalize
% Normalized
%The images are resized to 224×224 and processed through CLIP-ViT/32 to obtain 512-dimensional vectors and DINOv2-base to obtain 768-dimensional vectors.

%Let $\mathcal{D}_{\text{train}} = \{(x_i, y_i)\}_{i=1}^{N_{\text{train}}}$ denote the normalized training set embeddings with class labels, and $\mathcal{D}_{\text{test}} = \{(x_j, y_j)\}_{j=1}^{N_{\text{test}}}$ the normalized test set embeddings with their class labels. We define $\mathcal{C}$ as the set of class labels.

%change format
%\subsection{Two-stage procedure}
%local neighborhood selection: Why k-NN is replaced by NNK? 
%Local neighbors for the training set are obtained from the normalized latent space features of the foundation models, while neighbors are selected from the test set.
%\subsubsection{Estimating reliability from the training data} 

The high-level idea of our approach is illustrated in \autoref{fig:pipeline}. Given an FM, which we assume to be fixed, we represent all instances of a task in the embedding space produced by the FM. 
In the first stage (\autoref{sec:reliability}), we \textit{estimate reliability from the training data} by creating local neighborhoods for each training data embedding using the non-negative kernel (NNK) algorithm \cite{Shekkizhar2019}. The NNK algorithm constructs a sparse local neighborhood that includes only the geometrically non-redundant, most similar neighbors to the query, making it more suitable for this application than the k-NN method. 
Additionally, the NNK algorithm assigns normalized weights to the connections between the query and its neighbors based on the similarities in the local neighborhood. These local neighborhoods and NNK weights are then utilized for reliability score computation. 

In the second stage (\autoref{sec:inference}), 
%\subsubsection{Using reliability for inference on the test data} 
we \textit{use reliability for inference}. 
For each test data embedding, the local neighborhood is determined using the NNK algorithm, which includes only neighbors from the training data embeddings. Then, this neighborhood --along with the NNK weights if chosen-- is utilized for classification through weighted majority voting.
%The NNK algorithm \cite{Shekkizhar2019} generates a polytope around the query point using its nearest neighbors. It constructs a sparse local neighborhood consisting only of the most similar neighbors to the query, ensuring that these neighbors are distinct enough from one another. 

%The NNK algorithm generates two polytopes for each training data point in the embedding space. These polytopes are constructed from: 1) all neighboring training data points, and 2) those neighboring training data points that belong to the same class as the query.

%Additionally, the NNK algorithm assigns weights to the connections between the query and its neighbors based on the similarity between the query and each neighbor, as well as the similarity among the neighbors themselves. These weights indicate the degree of similarity between the query data and its neighbors, $w_{ix}$ indicating the weight between $i$ and the query $x$. 
%While the NNK algorithm constructs the polytopes, these associated weights are also stored for use in computing reliability metrics.

%BASELINE REL METRIC BASED ON p_e
\subsection{Reliability Metrics}
\label{sec:reliability}
% k-NN confidence: fraction of correct neighbors
% NNK diameter ratio (D/Dc): global vs same-class diameter
% Per-class geometric reliability: class-subset diameter -OMITTED
% Ensemble NNK reliability: average over subsamples
% WANN/ANN: 1/optimal-k reliability, weighted/ unweighted
% NNK-Means: inverse distance to same-label atoms
% k-Means: assigned centroids, binary

%\begin{itemize}
%\item \textbf{k-NN reliability}:
\subsubsection{k-NN reliability}
The reliability baseline score for each training data sample (the query $\xv_q$) is the fraction of $k$ nearest neighbors that share the same label as the query ($\hat y_q$): 
 \[ \eta_q=\frac{1}{k}\sum_{i=1}^{k}\mathbb{1}\bigl(\hat y_q = y_i\bigr). \]
%
%Skipping ANN and WANN
%\end{itemize}
%While k-NN reliability provides a simple measure of reliability based on the proportion of same‐class neighbors, it 
where $\mathbb{1}()$ is the indicator function. Note that this metric does not take into account the local geometry and density of the embedding space. In particular, it treats all neighbors equally. 
%and ignores how \textit{spread out} or \textit{tight} a local cluster is. 
%To overcome these limitations, we now introduce new geometry‐aware reliability metrics that explicitly leverage both the weights and the shape of each sample’s local neighborhood. Next, we define each metric in detail.
%
%\begin{itemize}
%\item \textbf{NNK weights reliability}: 
\subsubsection{Reliability based on NNK weights}
%When the embedding of the training data closely resembles that of its nearest neighbors with the same label, the reliability of the prediction increases, much like k-NN. 
%Unlike in k-NN reliability, here we consider the local distribution of points to estimate reliability. 
%However, the key difference with NNK is that it offers a sparser and geometrically non-redundant neighborhood.  
We propose to use the NNK local neighborhood construction \cite{Shekkizhar2019}, which, from any set of neighbors, can identify a subset that forms a polytope. 
In \cite{Shekkizhar2021Revisit}, a local interpolation was proposed, where the predicted label was a weighted function of the neighboring labels, with weights equal to the NNK neighborhood weights \( w_{iq} \).  
%The similarity between a query and its neighbors, which is based on Euclidean distances, is represented by the NNK weights, denoted as . 
Since the weights are normalized, we can quantify reliability from the NNK weights as the sum of the weights of local neighbors that share the same label with the query. 
%and then normalizes this sum by considering all neighbors. 
%Essentially, it is the ratio of the sum of the NNK weights for neighbors with the same labels to the total weights of all NNK neighbors. 
Denote \( N_q \) the NNK neighborhood of query \( \xv_q \) with label \( y_q \), and let \( y_i \) indicate the labels of its neighbors in  \( N_q \). 
We define the reliability score \( \hat \eta_q \) for $\xv_q$ as:
\begin{equation}
    \hat\eta_q={\underset{i \in N_{q}}{\sum } w_{iq} \mathbb{1} \left(y_{i}=y_{q}\right)},
    %\;\quad \underset{i \in N_{q}}{\sum } w_{iq}=1   
    \label{eq:NNKweightsrel}
\end{equation}
$\hat\eta_q = 1$ if all neighbors of $\xv_q$ have label $y_q$

%\item \textbf{NNK diameter ratio (D/Dc)}:
\subsubsection{NNK diameter ratio (D/Dc) reliability}
In \cite{Shekkizhar2021Revisit}, smaller NNK polytopes indicate less interpolation risk. Inspired by this idea, for each query training sample, we construct two distinct NNK polytopes: 1) one formed by all training neighbors, referred to as set \( S_q \), and 2) one formed only by neighbors that share the same class, \( \hat{S}_q \). We define the reliability score as the ratio of the diameters of these two polytopes, where the diameter of set \( S_q \), $\text{diam}(S_q)$, is defined as the maximum distance between neighbors $\xv_i\in S_q$. 
%of the query $\xv_q$. 
%For each data point in the training set and its known label $(\xv_q, y_q)$, and for its neighbors $(\xv_i, y_i)$ in set $S$, 
Thus for each query $\xv$ we define reliability as:
\begin{equation}
\hat\eta_q=\frac{\text{diam}(S_q)}{\text{diam}(\hat{S}_q)} \;\; \text{where} \;\; \hat{S}_q \subseteq{S_q}.
\end{equation}
Note that $\text{diam}(S_q) \leq \text{diam}(\hat{S}_q)$, since the average distances in $\hat{S}_q \geq S_q$. 
\subsubsection{Supervised k-means reliability }
As label noise increases, local neighborhoods can become less reliable (e.g., even if the label error probability is less than $0.5$, error can affect more than half the samples in a neighborhood), which can lead to incorrect majority vote results. 
This insight leads us to explore more global properties using cluster-based methods.
%We run standard k‐Means on the training embeddings to 
%We obtain 1-centroid per class on the training embeddings, $\mu_c$, for $j=\{1...C\}$.
%${\cv_j}$ for $j=\{1...C\}$, labeling each centroid by the mode of its cluster. Each sample $\xv_i$ is assigned to its nearest centroid index,
%\begin{equation*}
%m_i = \arg\min_j \lVert \xv_i - \cv_j\rVert_2,
%\end{equation*}
%and we set a binary reliability score based on the closest cluster center: 
%\begin{equation}
%\hat\eta_i= \mathbb{1} \bigl({y_{m_i} = y_i}\bigr)
%\in \{0,1\}
%\end{equation}
%supervised, 3*C clusters, 3-per class
%Using a single cluster centroid per class is insufficient for more complex data geometries. Instead of just one centroid, 
For each class, we run the k-means algorithm with \(K_c\) centroids:
\[
  \{\mu_{c,1},\dots,\mu_{c,K_c}\}
  \;=\;
  \mathrm{KMeans}\bigl(\{\mathbf{x}_i:y_i=c\},\,K_c\bigr),
\]
where 
\(\{\mu_j\}_{j=1}^M\) are the centroids, for \(M=K_c \times C\),  %% M=K_c* C, in experiments, K_c=3
and \(\mu_j\) has label \(\ell_j\in\{1,\dots,C\}\). For an arbitrary \emph{query sample} \(\mathbf{x}_q\), weights are computed \(\mathbf{w}_q = (w_{q,1},\dots,w_{q,M})\) as the softmax probabilities of distances:
\begin{equation}
w_{q,j} = \frac{\exp\bigl(-\,d(x_q,\mu_j)\bigr)}
       {\sum_{j'=1}^{M}\exp\bigl(-\,d(x_q,\mu_{j'})\bigr)}. \\
       \label{eq: softmax}
\end{equation}
Note that we base these weights on distances to all clusters across all labels. 
Then, given $y_q$, the reliability score is the maximum weight %(or the softmax distance for the closest centroid)
among all clusters in class $y_q$:
\begin{equation}
  \hat\eta_q
  \;=\;
  \max w_{q,j} \mathbb{1} \left(y_{q}=\ell_{j}\right).
\end{equation}
%and we set the reliability score based on the softmax probability of distance to the same-label centroid ($y_i=c$):
%\begin{equation}
%\hat\eta_i
%= \frac{1}{\lVert x_i - c_{j}\rVert_2 + \epsilon},
%\quad j \in \{1,\dots,C\},\quad
%\end{equation}
%where $y_{m_i}$ is the centroid’s label.

\subsubsection{Unsupervised k-means soft clustering reliability}  %unsupervised, 3*C clusters, probabilistic assignment
This method determines the cluster centers solely based on geometric principles. For each of the cluster centers, we assign a soft label based on the distribution of classes among samples assigned to the cluster. Unsupervised k-means with \(M\) clusters, (\(M \geq C\)),  is applied to obtain clusters, $\mathcal{K}_j$ and centroids, $\mu_j$:
\[
  \{\mathcal{K}_j\}_{j=1}^M
  = \mathrm{KMeans}\bigl(\{\mathbf{x}_i\};\,M\bigr),
  \quad
  \mu_j \;=\;\frac{1}{|\mathcal{K}_j|}\sum_{i\in\mathcal{K}_j}\mathbf{x}_i.
\]
%where \(\mathcal{K}_j = \{\,i : \text{cluster}(\mathbf{x}_i)=j\}\) and \(\mu_j\) is the centroid of cluster \(j\). 
Each centroid \(\mu_j\) is assigned a \emph{probabilistic label distribution}:
\begin{equation}
  p_j(c)
=\frac{\bigl|\{\,i\in\mathcal K_j: y_i=c\}\bigr|}{|\mathcal K_j|},\quad c=1,\dots,C.
  \label{eq: probdist}
\end{equation}
%%%OR
%\begin{equation}
%      p_{j}(c)
%      \;=\;
%      \frac{\#\{\,x_i\in\text{cluster }j : y_i = c\}}
%           {\#\{\,x_i\in\text{cluster }j\}},   \quad
%  c = 1,\dots,C.
%  \label{eq: probdist}
%\end{equation}

Given \(\mathbf{x}_q\), weights are computed \(\mathbf{w}_q = (w_{q,1},\dots,w_{q,M})\) as softmax of distances, as in \eqref{eq: softmax}. This k-means reliability-score combines soft cluster‑labels  \(p_j\) from \eqref{eq: probdist} and distance‑based weights \(w_{q,j}\) to quantify how confidently each sample supports its true class. Using the nearest cluster centroid, $\mu_{j*}$, when \(j^*= \arg\min \;d\bigl(x_q,\mu_{j*}\bigr)
\), and the reliability score $\hat\eta_q$  with query's known label \(y_q\) is
%change to nearest:
\begin{equation}
  \hat\eta_q
  \;=\; w_{q,j*}\;p_{j*}\bigl(y_q\bigr).
\end{equation}
%These reliability scores are used for inference using the NNK neighborhood, so weighted by $\hat\eta_i$ during inference.

%%CHANGED: WEIGHTED

%\item \textbf{NNK-Means}: %reliability = average inverse distance between dictionary atoms that share the same label.
\begin{comment}{
\subsubsection{NNK-means} 
In the k-means approach, the reliability score is derived from single cluster centers for each class, which can be affected by label noise. Using multiple cluster centers instead of just one can lead to more stable reliability scores. NNK-means\cite{shekkizhar2022nnkmeans} is a soft-clustering approach that identifies multiple clustering centers based on the NNK algorithm.
Using NNK-means, $m$ atoms are obtained for the training data $\xv_i$, atoms $\{a_{i1} \dots a_{im}\}$, serve as NNK cluster centers. 
Each atom has a distribution of labels , instead of a single label.
%majority class of its assigned points, $y^a_{ij}$. 
%For each training example $x_i$ with label $y_i$, we  compute the inverse Euclidean distance to its NNK cluster centers with the same label: %($y_i == y^a_{ij}$):
%\begin{equation}
%w_{iq}
%= \frac{1}{\lVert x_q - a_{i,j}\rVert_2 + \epsilon},
%\quad j=1,\dots,m,\quad    
%\end{equation}
Then define its reliability score $\hat\eta$ as
in \eqref{eq:NNKweightsrel}.
At test time, we use the same atom dictionary and ${\hat\eta}$ within our NNK‐based classifier, 
weighting each neighbor’s vote by its reliability.
%%CHANGED: distribution of classes : weighted assignment
}\end{comment}
%\end{itemize}
\subsection{Inference}
\label{sec:inference}

% Reliability-weighted voting: weight = reliability * edge weight or reliability alone
Predictions are generated using weighted majority voting, where each NNK neighbor's vote is adjusted according to its reliability. This adjustment has two options. In the weighted mode (W), the reliability is multiplied by the NNK edge weight, denoted as \( w_{it} \):
\begin{equation}
    \operatorname{W_{NNK weighted}}\left(x_{T}\right)=\underset{c \in C}{\arg \max } \sum_{i \in N_{T}} w_{it} \hat \eta_{i} \mathbb{1} \left(y_{i}=c\right).
     \label{eq:weighted}
\end{equation}
In the unweighted mode (UW), the vote is based solely on the reliability without any weighting 
\begin{equation}
    \operatorname{W_{NNK unweighted}}\left(x_{T}\right)=\underset{c \in C}{\arg \max } \sum_{i \in N_{T}} \hat \eta_{i} \mathbb{1} \left(y_{i}=c\right)
    \label{eq:unweighted}
\end{equation}
%\cite{Dudani1976weightmajvot,CoverHart1967majvot}
In \eqref{eq:weighted} and \eqref{eq:unweighted}, $\hat \eta_i$ represents the reliability score for each training data point $x_i$ within the local neighborhood $N_T$ of the test data $x_T$, $c$ denotes one of the classes in $C$.

\section{Simulation Results}
\subsection{Experimental Setup}
% Datasets: CIFAR-10, STL-10 ?, DermaMNIST
% Subsample 100/class, 
% Noise Types: Symmetrical, Assymmetrical
% Noise Levels: clean, 20%, 40%, 60% 
% NNK param: euclidean dist, k=50/100 , sigma=100 sqr(Dim)
% Reference methods: k-NN, ANN, WANN
% Protocol: Accuracy
We conduct experiments on two standard benchmark datasets,  under varying label noise conditions. We compare against three reference methods (k‐NN, ANN, WANN) and report classification accuracy on test samples.
\subsubsection{Datasets, model and preprocessing}
%To evaluate the effectiveness of our proposed reliability scores and classification framework in feature space, 
Following \cite{disalvo2025}, we use CIFAR-10 \cite{krizhevsky2009cifar10} subsampling 100 images per class and more challenging dataset, DermaMNIST \cite{medmnistv2}. All images are embedded using a fixed pre-trained network DINOv2 \cite{oquab2023dinov2}. 
The image data in both the training and test sets are resized to $224 \times 224$ using bilinear interpolation, resulting in 768-dimensional embedding vectors for the DINOv2-base model. These features are then L2-normalized.

\subsubsection{Noise protocol}
Following \cite{disalvo2025}, we inject symmetrical label noise at rates of $\{0\%,20\%,40\%,60\%\}$ and at rates of $\{0\%,20\%,30\%,40\%\}$ for asymmetrical label noise. Symmetric noise occurs when any label in a dataset is randomly switched with another label. In contrast, asymmetric noise involves a specific label being changed to a fixed label (for example, changing "bird" to "airplane"). Asymmetric noise simulates a systematic error, and it tends to group incorrect labels closely in the embedding space.
%We used the same protocol outlined in \cite{disalvo2025}.
\subsubsection{Baseline methods}
%(kmin, kmax) = (11, 51).
We sweep $k\in\{11,13,\dots,51\}$, for ANN and WANN, same as in \cite{disalvo2025}, and $k=50$ for k-NN .

\subsubsection{Reliability Configuration}
%NNK parameters and other hyperparameters
Our NNK method uses an initial neighbor set size of $k=50$ to initialize with k-NN and does not require a parameter sweep over $k$. We employ a Gaussian kernel with bandwidth 
\(
\sigma \;=\; 100\,\sqrt{d},
\)
where $d$ is the embedding dimension. In practice, we observe negligible performance variation if instead $\sigma=\sqrt{d}$. We noticed a minimal difference for $k > 50$. 
%needs some data / plots to support this claim
Euclidean distance serves as the similarity metric between the normalized data embeddings. 
%For the ensemble methods, the value of $R$ is set to 10, and the subsampling ratio is 50\%. %CHANGED in 5 RUNS 50%->80%
For supervised k-means, we select 1 centroid per class for  CIFAR-10 and 3 centroids per class for DermaMNIST. This choice was based on the increased complexity of the latter dataset. Further study of how to optimize the choice of cluster sizes is left for future work.  We use \( 3 \times C \) for unsupervised k-means, where \( C \) represents the number of classes. 
\subsubsection{Evaluation Metric}
All methods are evaluated by classification accuracy on the clean test set and clean and label-corrupted train set. We conduct our analysis over five runs, and we provide the mean and standard deviation of the results, summarized in the plots.

%\subsection{Geometric Evaluation Results}
% Radar plot/table comparing all FMs
% Results on polytope inclusion, residuals, sparsity
% FIG: Radar-plots?

\subsection{Comparison with Baselines}
% k-NN accuracy vs. NNK geometry
% TABLE: clustering metrics: k-NN Accuracy, Centroid Distance, Silhouette Coefficient
% When does geometry align or disagree?

%%PLOTS
\begin{figure*}[h]
    \hspace{2.5cm}
\includegraphics[width=0.85\linewidth]{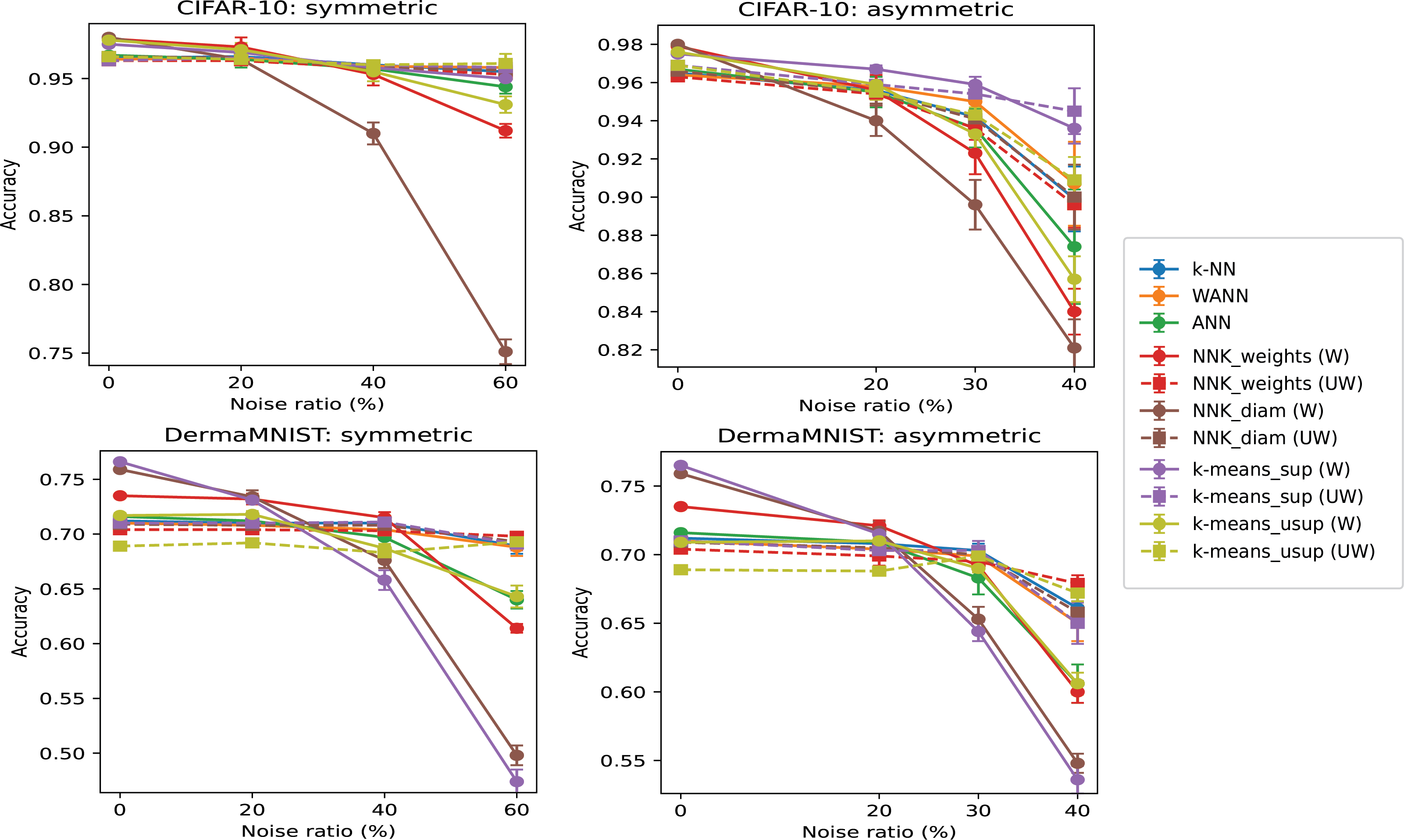}
  \caption{Accuracy vs. noise level (CIFAR-10 and DermaMNIST datasets) under symmetric and asymmetric label noise for 5 runs (mean $\pm$ std dev) across various reliability score methods and inference methods (weighted (W), unweighted (UW))}
    \label{fig:combined_FM}
\end{figure*}

\subsubsection{Local vs. global approaches} 
When most labels are correct, the precise distances in the embedding space are highly informative. Among our methods, the NNK‑diameter reliability score (weighted) depends on this local geometry the most and achieves the best accuracy on clean or lightly corrupted data. Traditional methods like WANN, ANN, and k‑NN also benefit from local data homogeneity but do not explicitly use distance weights; consequently, all NNK‑based scores outperform them under low noise (see \autoref{fig:combined_FM}).
As label noise increases, the benefit of purely local measures diminishes, so global methods based on clustering perform better. Cluster centers tend to shift less in the presence of label noise, even when the noise is systematic, as with asymmetric noise. Clustering methods (UW) demonstrate superior results for CIFAR-10 and DermaMNIST at increased noise levels, whether the noise is symmetric or asymmetric.
\subsubsection{Weighted vs. unweighted inference} At higher noise levels, unweighted majority voting tends to perform better, while weighted majority voting is more effective at lower noise levels. 
The performance gap between weighted and unweighted inference (shown as solid and dashed lines) widens in favor of unweighted inference at higher noise levels. 
This consistent trend indicates that distances in local neighborhoods become less trustworthy when many labels are wrong.
%This trend is evident in the simulation results for both CIFAR-10 and DermaMNIST across various noise types and levels, as shown in \autoref{fig:combined_FM}. 
\subsubsection{Accurate vs. inaccurate embeddings} The classification depends on how well the embedding provided by the FM captures class structure. When the task is relatively simple, as with CIFAR-10,  and FM embeddings are well matched to the task (higher accuracy), the local geometry proves to be useful, particularly when most labels are correct. However, with a harder dataset like DermaMNIST, we encounter noise in the geometry in addition to label noise. Thus, for DermaMNIST, at higher noise, unsupervised k-means (UW) and NNK weights (UW) reliabilities work the best.

%ENSEMBLE -NNK
%Improvements in robustness can be further enhanced using ensemble techniques. The ensemble NNK method helps to reduce the variance in reliability scores, making them more consistent. This is particularly beneficial in situations with higher label noise rates, where distances become less reliable and the local neighborhood can become more erroneous. In such cases, diversifying the neighborhood becomes essential.
%As shown in the plots, at higher noise rates, the geometry-based NNK weights, NNK diameter ratio, and NNK ensemble exhibit less variance compared to WANN, ANN, and k-NN. %-not so visible?
%Better performance for DermaMNIST

\subsubsection{Unsupervised vs. supervised k-means} For CIFAR-10, supervised k-means (W) outperforms its unsupervised counterpart for asymmetric label corruption. However, for DermaMNIST, at higher noise levels, unsupervised soft clustering k-means (UW) achieves better accuracy than supervised k-means. While supervised k-means (W) shows superior performance in low-noise scenarios, unsupervised k-means is more resilient to heavier noise in both symmetric and asymmetric cases. 
%This discrepancy is partly due to how well the class separation is defined in the embedding space. 
Geometry-only clustering (unsupervised) seems more helpful at high noise levels, where label errors can render per-label clusters meaningless. This effect is more significant for the more complex task, where geometric separation was not so good even without noise.  

\section{Conclusion}
% NNK-derived scores work well
% which one is better under which conditions
% geometric meaning and the potential of embedding space, explainability
%future work?
We propose and evaluate a variety of geometry-aware reliability estimators as part of a two-stage robust classification technique for foundation model embeddings under various noisy label scenarios that does not require model retraining. Our work emphasizes the effectiveness of geometry-based methods. 
%and leverages the plug-and-play nature of foundation models. 
Our findings indicate that at low noise levels, the geometry and distances in the embedding space are more important. 
%Consequently, reliability scores based on NNK methods, when weighted for NNK weights and reliability scores, outperform competitive approaches. 
As the noise level increases or for more complex embeddings, the effectiveness of distance-based measures diminishes. This can lead to local neighborhoods producing misleading reliability scores. Cluster-based methods address the problem by using global properties. 
%At higher noise levels, cluster-based methods outperform the competitive methods. 
Future work will explore adaptive hybrids that dynamically balance local and global methods, along with strategies for calibrating hyperparameters, such as neighborhood size and  number of clusters.

%comment on clustering methods
%We propose and evaluate a range of NNK-derived reliability estimators for foundational vision embeddings under different noisy-label scenarios, highlighting the effectiveness of NNK-based methods.

\vfill
\vfill

\pagebreak

\bibliographystyle{IEEEbib}
\bibliography{ref_foundationmodels}

\end{document}